\documentclass{article} % For LaTeX2e
\usepackage{iclr2023_conference_tinypaper,times}

\usepackage{amsmath,amsfonts,bm}

\def\eqref#1{equation~\ref{#1}}
\def\1{\bm{1}}

\DeclareMathAlphabet{\mathsfit}{\encodingdefault}{\sfdefault}{m}{sl}
\SetMathAlphabet{\mathsfit}{bold}{\encodingdefault}{\sfdefault}{bx}{n}

\usepackage{hyperref}
\usepackage{cleveref}
\usepackage{url}
\usepackage{booktabs}
\usepackage{overpic}
\usepackage{multicol}
\usepackage{multirow}
\usepackage[show]{chato-notes}
\title{Attention-likelihood relationship \\in transformers}

\author{
\parbox{\linewidth}{\centering
            Valeria Ruscio \thanks{Corresponding author: valeria.ruscio@uniroma1.it} \qquad
            Valentino Maiorca \qquad
            Fabrizio Silvestri \qquad 
 } \\[2ex]
{\parbox{\linewidth}{\centering
    Sapienza University of Rome 
      }
}
}

\iclrfinalcopy % Uncomment for camera-ready version, but NOT for submission.

\begin{document}

%\note{In my opinion it is not clear which is the impact of this paper. Having empirically proven what we write... what can be the impact on the literature or on the understanding of how LLMs work?}
\maketitle
\begin{abstract}
We analyze how large language models (LLMs) represent out-of-context words, investigating their reliance on the given context to capture their semantics. Our likelihood-guided text perturbations reveal a correlation between token likelihood and attention values in transformer-based language models. Extensive experiments reveal that unexpected tokens cause the model to attend less to the information coming from themselves to compute their representations, particularly at higher layers. These findings have valuable implications for assessing the robustness of LLMs in real-world scenarios. Fully reproducible codebase at \url{https://github.com/Flegyas/AttentionLikelihood}.

%force the model to rely more on context than the token itself to understand its meaning.

% In this study, we explore the connection between token likelihood and attention values in language models based on transformers. By perturbing individual tokens based on their likelihood, we observe that when unexpected tokens are introduced, the model tends to rely more on the contextual information to understand the semantics, as evidenced by an increase in attention values on contextual tokens and a decrease in attention values on the perturbed token.

\end{abstract}

\section{Introduction}
% \looseness=-1
% \cite{Kobayashi2020-ax}  propose a norm-based analysis that considers the norm of the transformed input vectors. 
% Transformers are the state-of-the-art class of language models introduced by \cite{Vaswani_undated-hp} that rely on self-attention to capture the contextual relationships between tokens in a sentence. While previous research has explored various aspects of the attention mechanism in transformers, such as specialized attention heads and the informative nature of attention weights \citep{Vig2019-mb}, \citep{Kovaleva2019-rg}, \citep{Clark2019-it}, other studies inquire whether the attention gives or not a full explanation on the transformer output \citep{Wiegreffe2019-cc}, \citep{Jain2019-ea}, \citep{Bibal_undated-bm}.

% However, much is still to learn about how this mechanism works in practice. 

% Towards this goal, we focus on the impact of the modeled probability distribution of a token (likelihood) on the self-attention mechanism in transformers. Our findings reveal that tokens with a higher likelihood value receive a correspondingly higher attention value, indicating that the model is relying on the token itself to understand its semantics. However, when out-of-context tokens are encountered, the model redirects its attention to the surrounding context, confirming that the attention mechanism in transformers is robust to new and unexpected situations. These insights can foster the development of more robust and accurate language models to enhance their ability to handle unexpected input. 

Transformers, introduced by \cite{Vaswani_undated-hp}, are the state-of-the-art architecture for language models that rely on self-attention to capture the contextual relationships between tokens in a sentence. While previous research has explored various aspects of the attention mechanism in transformers, such as specialized attention heads and the informative nature of attention weights \citep{Vig2019-mb, Kovaleva2019-rg, Clark2019-it}, other studies inquire whether attention alone can fully explain the transformer's output \citep{Wiegreffe2019-cc, Jain2019-ea, Bibal_undated-bm}.
Despite this previous work, there is still much to learn about the practical workings of the attention mechanism in transformers. In this study, we investigate the impact of token likelihood, i.e., the modeled probability distribution of a token, on the self-attention mechanism in transformers. Our findings show that tokens with a higher likelihood value receive a correspondingly higher attention value, indicating that the model is relying on the token itself to understand its semantics. However, when out-of-context (i.e., low-likelihood) tokens are encountered, the model redirects its attention to the surrounding context \textit{uniformly}, suggesting that the attention mechanism in transformers handles outliers by nullifying their information. These insights can foster the development of more robust and accurate language models, enhancing their ability to handle unexpected input.

\begin{figure}[h]
    \centering
    \begin{overpic}[trim=-0.27cm -0.15cm -.3cm 0cm,clip,width=0.9\linewidth]{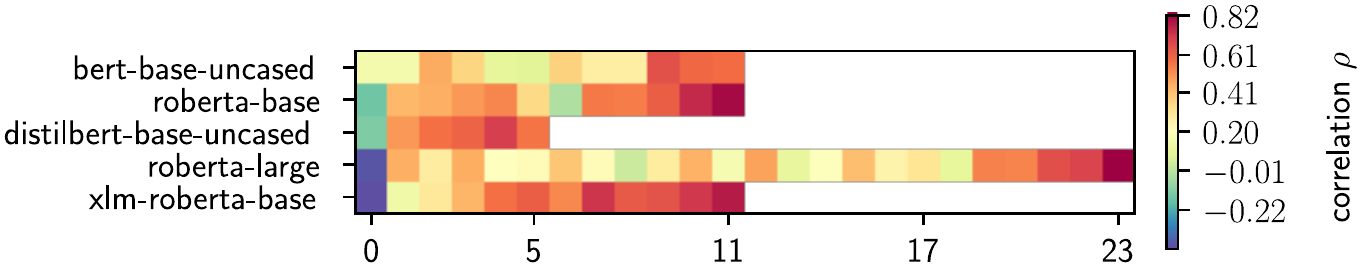}
    \end{overpic}
    \caption{The correlation $\rho$ between likelihood and token attention for various models and layers.}
    \label{fig:correlation}
\end{figure}

\section{Methodology}
\label{sec:method}
% \looseness=-1
We conduct a series of likelihood-guided perturbation experiments to investigate the impact of a token's likelihood on the self-attention mechanism in transformers. Let $M$ be a frozen autoencoder transformer model (e.g., BERT \citep{Devlin2018-uk}), and let $\mathcal{S}$ be a set of sentences. We are interested in two different outputs of $M$ applied to each $s \in \mathcal{S}$: \textbf{i)} the likelihood $L_{s_i}$ of token $s_i$ at position $i$ in $s$ as the conditional probability $P(s_i | s, M)$;  \textbf{ii)} the attention matrices  $\mathbf{A}^{(l)}$ for each multi-head attention layer $l$ of $M$. Note that we aggregate the outputs of the multiple attention heads in each layer into a single one via mean pooling.
After computing likelihood and attention on the original set $\mathcal{S}$, we formulate a likelihood-based perturbation on its sentences as follows: for each sentence $s \in S$, we select a token $s_i$ and replace it with a "perturbed" token $\hat{s_i}$, such that $L_{\hat{s}_i} \ll L_{s_i}$.
We then apply $M$ on this perturbed set $\hat{\mathcal{S}}$ to generate the perturbed likelihoods and attention matrices $\hat{\mathbf{A}}^(l)$.
We define the \textit{token attention} of the token $s_i$ at layer $l$ as $\mathbf{A}_{ii}^{(l)}$. Similarly, the \textit{sentence attention} is the concatenation of all the other attention weights on the $i$-th row of $\mathbf{A}$: $\oplus_{i\neq j}\mathbf{A}_{i,j}^{(l)}$. We compute them for both the original sentences $\mathcal{S}$ and the perturbed ones $\hat{\mathcal{S}}$.
% To analyze the impact of a token's likelihood on the self-attention mechanism, we compare the attention values of $\mathbf{A}_{ii}^{(l)}$ and $\hat{\mathbf{A}}_{ii}^{(l)}$ and perform a correlation analysis on the attention values of $\hat{\mathbf{A}}_{ii}^{(l)}$ to measure the relationship between the likelihood of a token $s_i$ and its corresponding attention weights.
% We repeat this process for each sentence in $\mathcal{S}$ to obtain a set of perturbed sentences and corresponding attention matrices for each transformer layer. We analyze the attention distribution across all tokens in each sentence to determine whether the model's attention shifts from the original token $s_i$ to the surrounding context when an unexpected token is encountered.
This methodology enables us to investigate the impact of a token's likelihood on the self-attention mechanism in transformers and probe the model's ability to adapt to unexpected input at its different layers while maintaining a controlled experimental setup by perturbing each sentence once and only one token per sentence.
%This study investigates the relationship between token likelihood and attention values in BERT-like \cite{Devlin2018-uk} transformers. Specifically, we compute the average attention value for a sentence, the average attention for a token with all the other tokens in the sentence (\textsc{sentence attention}), and the attention value for a token to the token itself (\textsc{token attention\footnote{we use "token attention" because "self-attention" would have been misleading.}}). Counter-intuitively, we find that the inner attention value is significantly higher than the outer attention value, suggesting that the most relevant word to define a token is often the word itself. 
%\inote{Here, we are missing the conclusions we can draw from this observation.}

%We perform likelihood-guided perturbations (further explanations in appendix) on individual tokens in the sentence to further explore this relationship. We switch some of the token's sub-word units (i.e., wordpieces as in \cite{Schuster2012-ro}) with other wordpieces with a low \inote{lower?} likelihood value and measure the resulting changes in attention values. 

%We find that when we perturb the token, the inner attention value decreases significantly, while the outer attention value for the sentence increases, suggesting that an unexpected token causes the model to rely more on the context than the token to grasp its semantics. 

\section{Experiments}
\looseness-1

\begin{table}[]
\centering
\small

\caption{The table shows, for both datasets, the mean: token attention $\mathbf{A_{ii}}$, correlation $\rho$ and likelihood $L_{s}$. The Mann-Whitney U (MWU) is tested between the sentence attentions of the two.}

\begin{tabular}{ccrrrrrrrr}
\toprule
                        &       & \multicolumn{3}{c}{\textbf{Original} ($\mathcal{S}$)}               & \multicolumn{3}{c}{\textbf{Perturbed}  ($\mathcal{\hat{S}}$)} &        \textbf{MWU }                  \\ 
\cmidrule(lr){3-5}\cmidrule(lr){6-8}\cmidrule(l){9-9}
\textbf{encoder}                 & \textbf{layer} & $\bar{\mathbf{A_{ii}}}$ & $\bar{L_{s}}$    & $\bar{\rho}$  & $\bar{\mathbf{\hat{A_{ii}}}}$  & $\bar{\hat{L_{s}}}$    & $\bar{\hat{\rho}}$ &\textbf{ p-value} \\
\toprule
BERT    & 9        & 0.075  & 0.664 & 0.095       &  0.063  & 0.096 & 0.643  & 0.52     \\
RoBERTa base            & 11         & 0.124  & 0.938 & 0.247         & 0.043  & 0.072 & 0.802     & 0.54  \\
XLM-R        & 11        & 0.132  & 0.943 & 0.332         & 0.051 & 0.017 & 0.767    & 0.48    \\
RoBERTa large           & 23         & 0.134 & 0.947 & 0.183         & 0.046  & 0.055 & 0.823   & 0.56    \\
DistilBERT & 4         & 0.065  & 0.785 & 0.033          & 0.045 & 0.017 & 0.688  & 0.55 \\
\bottomrule
\label{table:correlation}
\end{tabular}
\end{table}
\label{sec:experiments}
We conduct experiments on a dataset $\mathcal{S}$ consisting of around 20k English sentences extracted from WordNet \citep{Miller1995-os}. Our analysis involves five transformer models, namely RoBERTa \citep{Liu2019-fq} (\textit{base} and \textit{large} versions), BERT \citep{Devlin2018-uk} and DistilBERT \citep{Sanh2019-lo} (\textit{base uncased} versions), and XLM-R \citep{xlmr}. To investigate the relationship between token likelihood and token attention, we use the Spearman correlation metric on both the original and the perturbed dataset. We report the results of our experiments in \cref{table:correlation}. The complete version can be found in the appendix along with additional dataset details.
Our findings reveal a \textbf{strong correlation} between token likelihood and attention values. Specifically, the token's attention decreases significantly when its likelihood is lowered, indicating that when the model encounters an unexpected token, it shifts its focus from the token to the context to compute its representation. We observe that the attention previously directed to the token is then distributed uniformly to the rest of the sentence, as confirmed by the results of the Mann-Whitney U test \citep{McKnight2010-wx} applied between the attention distributions.
Furthermore, we observe that the attention changes due to likelihood-guided perturbations are relevant across most model layers. However, the correlation values are stronger in the latest layers that encode more abstract semantic features \cite{Rogers_undated-it, Liu2019-vd}. In contrast, in the first layers, the model relies more on the token itself to understand its meaning \citep{Vig2019-mb}, even when perturbed.

\section{Conclusion}
% \looseness-1
Our study thoroughly analyzed the link between token likelihood and the self-attention mechanism in transformers, finding a strong statistical correlation between them. This can be seen as a transformer strategy to handle out-of-context words since it redirects its focus from the unlikely token to the context to grasp its semantics, especially at deeper attention layers.
Future research could explore the implications of these findings on downstream tasks, enhance language models' robustness, and investigate this phenomenon in different domains, such as ViTs \citep{vit}. This could further our understanding of the self-attention mechanism and its application to other fields. 

\subsubsection*{Acknowledgements}
This work is supported by the ERC Starting Grant No. 802554 (SPECGEO).

% \subsection*{URM Statement}
% The authors acknowledge that at least one key author of this work meets the URM criteria of ICLR 2023 Tiny Papers Track.
%Please include this URM Statement section at the end of the paper but before the references before. In your anonymized submission, we recommend stating ``The authors acknowledge that at least one key author of this work meets the URM criteria of ICLR 2023 Tiny Papers Track.'' For the camera ready version, we ask authors to identify which author(s) meet the URM criteria, e.g., ``Author TFB meets the URM criteria of ICLR 2023 Tiny Papers Track.'' The authors are also welcome to come up with their own phrases to affirm meeting this criteria.

\bibliography{iclr2023_conference_tinypaper}
\bibliographystyle{iclr2023_conference_tinypaper}

\appendix
\section{Appendix}
\subsection{Implementation Details}
We follow the standard best practices for our codebase, both for deep learning and for Python development in general. These are the core tools we relied on:
\begin{itemize}
    \item \textit{Transformers by HuggingFace} \citep{transformers}, to get ready-to-use transformer-based LLMs;
    \item \textit{Datasets by HuggingFace} \citep{datasets}, to handle the creation and processing of our dataset;
    \item \textit{DVC} \citep{dvc}, for versioning our datasets and results accordingly;
    \item \textit{NLTK} \citep{nltk} for easy access to WordNet.
\end{itemize}

\begin{table}[h]
\centering
\caption{The HuggingFace transformers used in \Cref{sec:experiments}, along with some of their specifics.}
\label{tab:transformers-nlp}
\begin{tabular}{crr}
\toprule
official name    & Num. Layers  & Num. Params \\
\midrule
roberta-base                     & 12     & 125M     \\
bert-base-uncased   & 12 & 110M        \\
roberta-large     & 24  & 355M        \\
xlm-roberta-base & 12   & 125M       \\
distilbert-base-uncased & 6 & 66M          \\
\bottomrule
\end{tabular}
\end{table}

\subsection{Data}
\label{extra:data}

To produce the dataset we utilized WordNet. For each synset in it, we retrieved all the lemmas and all the examples. Then, we kept only the examples having an exact match with at least one of the lemmas for their synset. We tokenized the remaining sentences and discarded the ones shorter than 5 tokens. With this procedure, we ended up with a total of 20285 sentences.

This was done to ensure both high-quality sentences (WordNet is manually annotated) and that the target token for the perturbation is a "meaningful" one and not a stopword (which could've added negative biases).

\begin{figure}[h]
    \centering
    \begin{overpic}[trim=-0.27cm -0.15cm -.3cm 0cm,clip,width=0.9\linewidth]{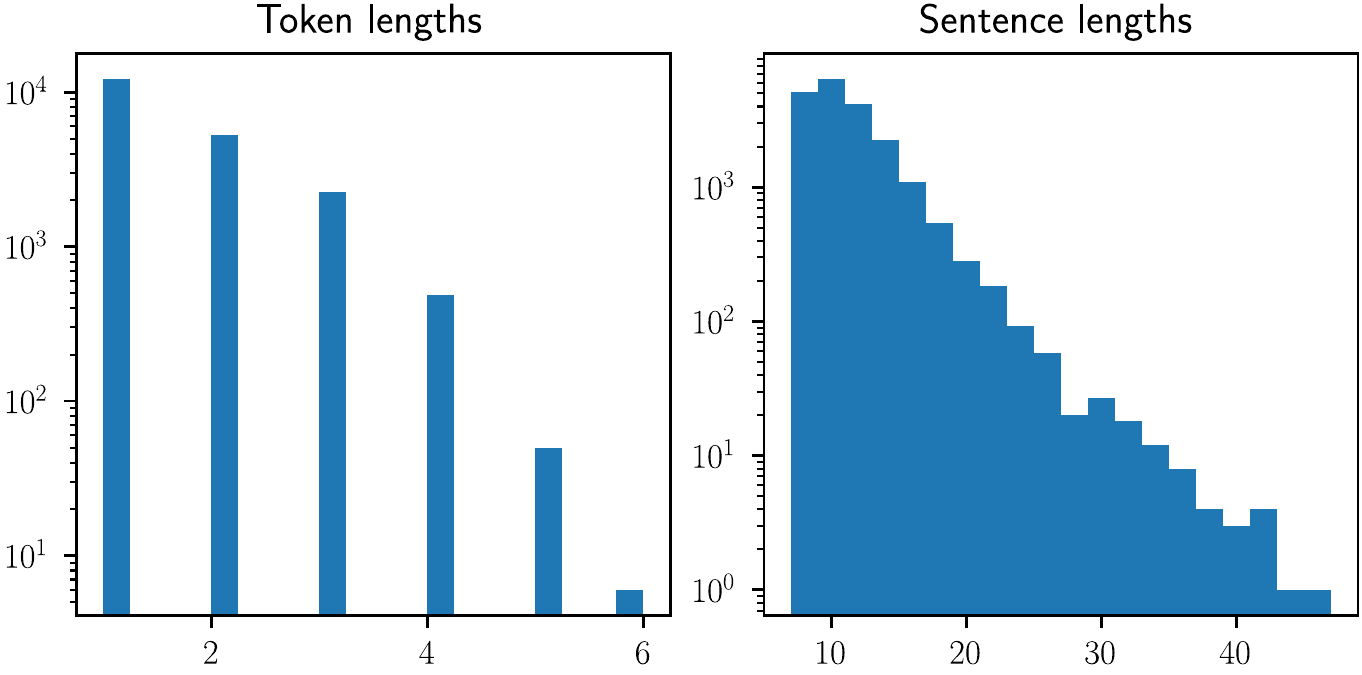}
    \end{overpic}
    \caption{The distributions of the perturbed token lengths (\textit{left}) and the sentence lengths (\textit{right}). Both are calculated in number of wordpieces.}
    \label{fig:correlation}
\end{figure}

\subsection{Full Results}

\begin{table}[h]
\scriptsize
\caption{The table shows, for the original dataset $\mathcal{S}$, the mean $\pm$ std of: attention $\mathbf{A}$, token attention $\mathbf{A_{ii}}$, sentence attention $\mathbf{A_{ij}}$, correlation $\rho$ and likelihood $L_{s}$.}
\centering
\begin{tabular}{ccrrrrr}
\toprule
\textbf{encoder}                                  & \textbf{layer} & $\bar{\mathbf{A}}$     & $\bar{\mathbf{A_{ii}}}$ & $\bar{\mathbf{A_{ij}}}$ & $\bar{L_{s}}$ & $\bar{\rho}$ \\ \midrule
\multirow{12}{*}{BERT}      & 0     & 0.099 ± 0.026 & 0.110 ± 0.028    & 0.098 ± 0.026    & 0.664 ± 0.356     & -0.146      \\
                                         & 1     & 0.099 ± 0.026 & 0.062 ± 0.018    & 0.100 ± 0.026    & 0.664 ± 0.356     & 0.008       \\
                                         & 2     & 0.099 ± 0.026 & 0.078 ± 0.022    & 0.099 ± 0.026    & 0.664 ± 0.356     & -0.035      \\
                                         & 3     & 0.099 ± 0.026 & 0.047 ± 0.021    & 0.100 ± 0.026    & 0.664 ± 0.356     & -0.220      \\
                                         & 4     & 0.099 ± 0.026 & 0.056 ± 0.023    & 0.100 ± 0.026    & 0.664 ± 0.356     & -0.236      \\
                                         & 5     & 0.099 ± 0.026 & 0.034 ± 0.016    & 0.101 ± 0.026    & 0.664 ± 0.356     & -0.291      \\
                                         & 6     & 0.099 ± 0.026 & 0.040 ± 0.018    & 0.100 ± 0.026    & 0.664 ± 0.356     & -0.222      \\
                                         & 7     & 0.099 ± 0.026 & 0.038 ± 0.018    & 0.101 ± 0.026    & 0.664 ± 0.356     & -0.314      \\
                                         & 8     & 0.099 ± 0.026 & 0.038 ± 0.017    & 0.101 ± 0.026    & 0.664 ± 0.356     & -0.150      \\
                                         & 9     & 0.099 ± 0.026 & 0.075 ± 0.032    & 0.100 ± 0.026    & 0.664 ± 0.356     & 0.095       \\
                                         & 10    & 0.099 ± 0.026 & 0.089 ± 0.049    & 0.100 ± 0.026    & 0.664 ± 0.356     & 0.163       \\
                                         & 11    & 0.099 ± 0.026 & 0.070 ± 0.030    & 0.100 ± 0.026    & 0.664 ± 0.356     & 0.266       \\
\midrule
\multirow{12}{*}{RoBERTa base}           & 0     & 0.100 ± 0.025 & 0.158 ± 0.031    & 0.098 ± 0.025    & 0.938 ± 0.197     & 0.015       \\
                                         & 1     & 0.100 ± 0.025 & 0.052 ± 0.029    & 0.101 ± 0.026    & 0.938 ± 0.197     & 0.027       \\
                                         & 2     & 0.100 ± 0.025 & 0.057 ± 0.019    & 0.101 ± 0.026    & 0.938 ± 0.197     & 0.077       \\
                                         & 3     & 0.100 ± 0.025 & 0.047 ± 0.015    & 0.101 ± 0.026    & 0.938 ± 0.197     & 0.064       \\
                                         & 4     & 0.100 ± 0.025 & 0.052 ± 0.018    & 0.101 ± 0.025    & 0.938 ± 0.197     & -0.190      \\
                                         & 5     & 0.100 ± 0.025 & 0.048 ± 0.017    & 0.101 ± 0.025    & 0.938 ± 0.197     & -0.014      \\
                                         & 6     & 0.100 ± 0.025 & 0.052 ± 0.022    & 0.101 ± 0.025    & 0.938 ± 0.197     & -0.125      \\
                                         & 7     & 0.100 ± 0.025 & 0.062 ± 0.024    & 0.100 ± 0.026    & 0.938 ± 0.197     & -0.072      \\
                                         & 8     & 0.100 ± 0.025 & 0.045 ± 0.021    & 0.101 ± 0.026    & 0.938 ± 0.197     & -0.246      \\
                                         & 9     & 0.100 ± 0.025 & 0.054 ± 0.035    & 0.101 ± 0.026    & 0.938 ± 0.197     & -0.275      \\
                                         & 10    & 0.100 ± 0.025 & 0.066 ± 0.027    & 0.100 ± 0.025    & 0.938 ± 0.197     & -0.245      \\
                                         & 11    & 0.100 ± 0.025 & 0.124 ± 0.048    & 0.099 ± 0.025    & 0.938 ± 0.197     & 0.247       \\
\midrule
\multirow{12}{*}{XLM-R}       & 0     & 0.089 ± 0.024 & 0.123 ± 0.039    & 0.089 ± 0.024    & 0.943 ± 0.176     & -0.088      \\
                                         & 1     & 0.089 ± 0.024 & 0.071 ± 0.031    & 0.090 ± 0.025    & 0.943 ± 0.176     & -0.076      \\
                                         & 2     & 0.089 ± 0.024 & 0.058 ± 0.023    & 0.090 ± 0.025    & 0.943 ± 0.176     & 0.122       \\
                                         & 3     & 0.089 ± 0.024 & 0.075 ± 0.020    & 0.090 ± 0.025    & 0.943 ± 0.176     & 0.130       \\
                                         & 4     & 0.089 ± 0.024 & 0.061 ± 0.018    & 0.090 ± 0.025    & 0.943 ± 0.176     & 0.075       \\
                                         & 5     & 0.089 ± 0.024 & 0.068 ± 0.014    & 0.090 ± 0.025    & 0.943 ± 0.176     & -0.258      \\
                                         & 6     & 0.089 ± 0.024 & 0.045 ± 0.017    & 0.091 ± 0.025    & 0.943 ± 0.176     & 0.002       \\
                                         & 7     & 0.089 ± 0.024 & 0.054 ± 0.017    & 0.091 ± 0.025    & 0.943 ± 0.176     & -0.548      \\
                                         & 8     & 0.089 ± 0.024 & 0.055 ± 0.019    & 0.091 ± 0.025    & 0.943 ± 0.176     & -0.407      \\
                                         & 9     & 0.089 ± 0.024 & 0.083 ± 0.030    & 0.090 ± 0.024    & 0.943 ± 0.176     & -0.441      \\
                                         & 10    & 0.089 ± 0.024 & 0.082 ± 0.037    & 0.090 ± 0.024    & 0.943 ± 0.176     & -0.543      \\
                                         & 11    & 0.089 ± 0.024 & 0.132 ± 0.063    & 0.089 ± 0.024    & 0.943 ± 0.176     & 0.332       \\
\midrule
\multirow{24}{*}{RoBERTa large}          & 0     & 0.100 ± 0.025 & 0.199 ± 0.073    & 0.098 ± 0.024    & 0.947 ± 0.172     & -0.166      \\
                                         & 1     & 0.100 ± 0.025 & 0.047 ± 0.030    & 0.101 ± 0.026    & 0.947 ± 0.172     & 0.180       \\
                                         & 2     & 0.100 ± 0.025 & 0.041 ± 0.024    & 0.101 ± 0.026    & 0.947 ± 0.172     & 0.043       \\
                                         & 3     & 0.100 ± 0.025 & 0.046 ± 0.023    & 0.101 ± 0.026    & 0.947 ± 0.172     & 0.088       \\
                                         & 4     & 0.100 ± 0.025 & 0.053 ± 0.019    & 0.101 ± 0.025    & 0.947 ± 0.172     & -0.082      \\
                                         & 5     & 0.100 ± 0.025 & 0.026 ± 0.015    & 0.102 ± 0.026    & 0.947 ± 0.172     & 0.008       \\
                                         & 6     & 0.100 ± 0.025 & 0.029 ± 0.012    & 0.102 ± 0.026    & 0.947 ± 0.172     & 0.009       \\
                                         & 7     & 0.100 ± 0.025 & 0.062 ± 0.020    & 0.101 ± 0.025    & 0.947 ± 0.172     & -0.189      \\
                                         & 8     & 0.100 ± 0.025 & 0.079 ± 0.025    & 0.101 ± 0.025    & 0.947 ± 0.172     & -0.164      \\
                                         & 9     & 0.100 ± 0.025 & 0.057 ± 0.018    & 0.101 ± 0.026    & 0.947 ± 0.172     & -0.121      \\
                                         & 10    & 0.100 ± 0.025 & 0.053 ± 0.017    & 0.101 ± 0.025    & 0.947 ± 0.172     & -0.113      \\
                                         & 11    & 0.100 ± 0.025 & 0.052 ± 0.020    & 0.101 ± 0.025    & 0.947 ± 0.172     & -0.032      \\
                                         & 12    & 0.100 ± 0.025 & 0.052 ± 0.017    & 0.101 ± 0.025    & 0.947 ± 0.172     & -0.154      \\
                                         & 13    & 0.100 ± 0.025 & 0.042 ± 0.021    & 0.101 ± 0.025    & 0.947 ± 0.172     & -0.048      \\
                                         & 14    & 0.100 ± 0.025 & 0.032 ± 0.018    & 0.102 ± 0.026    & 0.947 ± 0.172     & -0.186      \\
                                         & 15    & 0.100 ± 0.025 & 0.040 ± 0.019    & 0.101 ± 0.026    & 0.947 ± 0.172     & -0.020      \\
                                         & 16    & 0.100 ± 0.025 & 0.042 ± 0.018    & 0.101 ± 0.026    & 0.947 ± 0.172     & -0.061      \\
                                         & 17    & 0.100 ± 0.025 & 0.042 ± 0.019    & 0.101 ± 0.026    & 0.947 ± 0.172     & -0.013      \\
                                         & 18    & 0.100 ± 0.025 & 0.038 ± 0.019    & 0.102 ± 0.026    & 0.947 ± 0.172     & -0.112      \\
                                         & 19    & 0.100 ± 0.025 & 0.045 ± 0.013    & 0.101 ± 0.026    & 0.947 ± 0.172     & -0.073      \\
                                         & 20    & 0.100 ± 0.025 & 0.055 ± 0.019    & 0.101 ± 0.026    & 0.947 ± 0.172     & -0.007      \\
                                         & 21    & 0.100 ± 0.025 & 0.068 ± 0.034    & 0.100 ± 0.026    & 0.947 ± 0.172     & -0.195      \\
                                         & 22    & 0.100 ± 0.025 & 0.052 ± 0.030    & 0.101 ± 0.026    & 0.947 ± 0.172     & -0.171      \\
                                         & 23    & 0.100 ± 0.025 & 0.134 ± 0.040    & 0.099 ± 0.025    & 0.947 ± 0.172     & 0.183       \\
\midrule
\multirow{6}{*}{DistilBERT} & 0     & 0.099 ± 0.026 & 0.096 ± 0.023    & 0.099 ± 0.026    & 0.785 ± 0.293     & -0.202      \\
                                         & 1     & 0.099 ± 0.026 & 0.091 ± 0.020    & 0.099 ± 0.026    & 0.785 ± 0.293     & -0.111      \\
                                         & 2     & 0.099 ± 0.026 & 0.045 ± 0.017    & 0.100 ± 0.026    & 0.785 ± 0.293     & -0.163      \\
                                         & 3     & 0.099 ± 0.026 & 0.039 ± 0.018    & 0.100 ± 0.026    & 0.785 ± 0.293     & -0.185      \\
                                         & 4     & 0.099 ± 0.026 & 0.065 ± 0.022    & 0.100 ± 0.026    & 0.785 ± 0.293     & 0.033       \\
                                         & 5     & 0.099 ± 0.026 & 0.063 ± 0.024    & 0.100 ± 0.026    & 0.785 ± 0.293     & 0.017       \\ \bottomrule
\end{tabular}
\end{table}

\begin{table}[ht]
\scriptsize
\caption{The table shows, for the perturbed dataset $\mathcal{\hat{S}}$, the mean $\pm$ std of: attention $\mathbf{\hat{A}}$, token attention $\mathbf{\hat{A}_{ii}}$, sentence attention $\mathbf{\hat{A}_{ij}}$, correlation $\hat{\rho}$ and likelihood $\hat{L}_{s}$.}
\centering
\begin{tabular}{ccrrrrr}
\toprule
\textbf{encoder}                                  & \textbf{layer} & $\bar{\mathbf{A}}$     & $\bar{\mathbf{A_{ii}}}$ & $\bar{\mathbf{A_{ij}}}$ & $\bar{L_{s}}$ & $\bar{\rho}$ \\ \midrule
\multirow{12}{*}{BERT}      & 0     & 0.099 ± 0.026 & 0.099 ± 0.027    & 0.099 ± 0.026    & 0.096 ± 0.198     & 0.143       \\
                                         & 1     & 0.099 ± 0.026 & 0.050 ± 0.017    & 0.100 ± 0.026    & 0.096 ± 0.198     & 0.142       \\
                                         & 2     & 0.099 ± 0.026 & 0.081 ± 0.023    & 0.099 ± 0.026    & 0.096 ± 0.198     & 0.447       \\
                                         & 3     & 0.099 ± 0.026 & 0.049 ± 0.022    & 0.100 ± 0.026    & 0.096 ± 0.198     & 0.345       \\
                                         & 4     & 0.099 ± 0.026 & 0.065 ± 0.021    & 0.100 ± 0.026    & 0.096 ± 0.198     & 0.084       \\
                                         & 5     & 0.099 ± 0.026 & 0.051 ± 0.023    & 0.100 ± 0.026    & 0.096 ± 0.198     & 0.065       \\
                                         & 6     & 0.099 ± 0.026 & 0.049 ± 0.023    & 0.100 ± 0.026    & 0.096 ± 0.198     & 0.353       \\
                                         & 7     & 0.099 ± 0.026 & 0.060 ± 0.035    & 0.100 ± 0.026    & 0.096 ± 0.198     & 0.264       \\
                                         & 8     & 0.099 ± 0.026 & 0.047 ± 0.021    & 0.100 ± 0.026    & 0.096 ± 0.198     & 0.262       \\
                                         & 9     & 0.099 ± 0.026 & 0.063 ± 0.041    & 0.100 ± 0.026    & 0.096 ± 0.198     & 0.643       \\
                                         & 10    & 0.099 ± 0.026 & 0.062 ± 0.052    & 0.100 ± 0.026    & 0.096 ± 0.198     & 0.588       \\
                                         & 11    & 0.099 ± 0.026 & 0.045 ± 0.035    & 0.100 ± 0.026    & 0.096 ± 0.198     & 0.576       \\
                                         \midrule
\multirow{12}{*}{RoBERTa base}           & 0     & 0.100 ± 0.025 & 0.166 ± 0.039    & 0.098 ± 0.025    & 0.072 ± 0.174     & -0.157      \\
                                         & 1     & 0.100 ± 0.025 & 0.050 ± 0.023    & 0.101 ± 0.026    & 0.072 ± 0.174     & 0.417       \\
                                         & 2     & 0.100 ± 0.025 & 0.050 ± 0.018    & 0.101 ± 0.026    & 0.072 ± 0.174     & 0.440       \\
                                         & 3     & 0.100 ± 0.025 & 0.038 ± 0.018    & 0.101 ± 0.026    & 0.072 ± 0.174     & 0.488       \\
                                         & 4     & 0.100 ± 0.025 & 0.049 ± 0.021    & 0.101 ± 0.026    & 0.072 ± 0.174     & 0.521       \\
                                         & 5     & 0.100 ± 0.025 & 0.038 ± 0.015    & 0.101 ± 0.026    & 0.072 ± 0.174     & 0.337       \\
                                         & 6     & 0.100 ± 0.025 & 0.064 ± 0.019    & 0.101 ± 0.025    & 0.072 ± 0.174     & -0.038      \\
                                         & 7     & 0.100 ± 0.025 & 0.062 ± 0.028    & 0.100 ± 0.026    & 0.072 ± 0.174     & 0.544       \\
                                         & 8     & 0.100 ± 0.025 & 0.057 ± 0.035    & 0.100 ± 0.026    & 0.072 ± 0.174     & 0.541       \\
                                         & 9     & 0.100 ± 0.025 & 0.073 ± 0.051    & 0.100 ± 0.026    & 0.072 ± 0.174     & 0.609       \\
                                         & 10    & 0.100 ± 0.025 & 0.071 ± 0.048    & 0.100 ± 0.026    & 0.072 ± 0.174     & 0.739       \\
                                         & 11    & 0.100 ± 0.025 & 0.043 ± 0.033    & 0.101 ± 0.026    & 0.072 ± 0.174     & 0.802       \\
                                         \midrule
\multirow{12}{*}{XLM-R}       & 0     & 0.089 ± 0.024 & 0.151 ± 0.053    & 0.088 ± 0.024    & 0.017 ± 0.071     & -0.429      \\
                                         & 1     & 0.089 ± 0.024 & 0.116 ± 0.031    & 0.088 ± 0.024    & 0.017 ± 0.071     & 0.125       \\
                                         & 2     & 0.089 ± 0.024 & 0.065 ± 0.022    & 0.090 ± 0.025    & 0.017 ± 0.071     & 0.283       \\
                                         & 3     & 0.089 ± 0.024 & 0.059 ± 0.020    & 0.090 ± 0.025    & 0.017 ± 0.071     & 0.425       \\
                                         & 4     & 0.089 ± 0.024 & 0.058 ± 0.022    & 0.090 ± 0.025    & 0.017 ± 0.071     & 0.565       \\
                                         & 5     & 0.089 ± 0.024 & 0.058 ± 0.017    & 0.090 ± 0.025    & 0.017 ± 0.071     & 0.610       \\
                                         & 6     & 0.089 ± 0.024 & 0.051 ± 0.021    & 0.091 ± 0.025    & 0.017 ± 0.071     & 0.516       \\
                                         & 7     & 0.089 ± 0.024 & 0.055 ± 0.023    & 0.090 ± 0.025    & 0.017 ± 0.071     & 0.713       \\
                                         & 8     & 0.089 ± 0.024 & 0.057 ± 0.026    & 0.090 ± 0.025    & 0.017 ± 0.071     & 0.622       \\
                                         & 9     & 0.089 ± 0.024 & 0.095 ± 0.042    & 0.089 ± 0.025    & 0.017 ± 0.071     & 0.638       \\
                                         & 10    & 0.089 ± 0.024 & 0.083 ± 0.050    & 0.089 ± 0.025    & 0.017 ± 0.071     & 0.709       \\
                                         & 11    & 0.089 ± 0.024 & 0.051 ± 0.034    & 0.090 ± 0.025    & 0.017 ± 0.071     & 0.767       \\ 
                                         \midrule
\multirow{24}{*}{RoBERTa large}          & 0     & 0.100 ± 0.025 & 0.219 ± 0.075    & 0.097 ± 0.024    & 0.055 ± 0.149     & -0.410      \\
                                         & 1     & 0.100 ± 0.025 & 0.034 ± 0.020    & 0.101 ± 0.026    & 0.055 ± 0.149     & 0.439       \\
                                         & 2     & 0.100 ± 0.025 & 0.036 ± 0.020    & 0.101 ± 0.026    & 0.055 ± 0.149     & 0.268       \\
                                         & 3     & 0.100 ± 0.025 & 0.041 ± 0.018    & 0.101 ± 0.026    & 0.055 ± 0.149     & 0.441       \\
                                         & 4     & 0.100 ± 0.025 & 0.048 ± 0.019    & 0.101 ± 0.026    & 0.055 ± 0.149     & 0.195       \\
                                         & 5     & 0.100 ± 0.025 & 0.021 ± 0.013    & 0.102 ± 0.026    & 0.055 ± 0.149     & 0.209       \\
                                         & 6     & 0.100 ± 0.025 & 0.031 ± 0.014    & 0.101 ± 0.026    & 0.055 ± 0.149     & 0.387       \\
                                         & 7     & 0.100 ± 0.025 & 0.058 ± 0.022    & 0.101 ± 0.025    & 0.055 ± 0.149     & 0.214       \\
                                         & 8     & 0.100 ± 0.025 & 0.074 ± 0.026    & 0.100 ± 0.025    & 0.055 ± 0.149     & 0.018       \\
                                         & 9     & 0.100 ± 0.025 & 0.050 ± 0.018    & 0.101 ± 0.026    & 0.055 ± 0.149     & 0.271       \\
                                         & 10    & 0.100 ± 0.025 & 0.048 ± 0.015    & 0.101 ± 0.026    & 0.055 ± 0.149     & 0.432       \\
                                         & 11    & 0.100 ± 0.025 & 0.040 ± 0.017    & 0.101 ± 0.026    & 0.055 ± 0.149     & 0.156       \\
                                         & 12    & 0.100 ± 0.025 & 0.052 ± 0.016    & 0.101 ± 0.026    & 0.055 ± 0.149     & 0.470       \\
                                         & 13    & 0.100 ± 0.025 & 0.031 ± 0.014    & 0.102 ± 0.026    & 0.055 ± 0.149     & 0.093       \\
                                         & 14    & 0.100 ± 0.025 & 0.030 ± 0.012    & 0.102 ± 0.026    & 0.055 ± 0.149     & 0.189       \\
                                         & 15    & 0.100 ± 0.025 & 0.030 ± 0.012    & 0.102 ± 0.026    & 0.055 ± 0.149     & 0.407       \\
                                         & 16    & 0.100 ± 0.025 & 0.037 ± 0.013    & 0.101 ± 0.026    & 0.055 ± 0.149     & 0.244       \\
                                         & 17    & 0.100 ± 0.025 & 0.041 ± 0.014    & 0.101 ± 0.026    & 0.055 ± 0.149     & 0.299       \\
                                         & 18    & 0.100 ± 0.025 & 0.042 ± 0.017    & 0.101 ± 0.026    & 0.055 ± 0.149     & 0.088       \\
                                         & 19    & 0.100 ± 0.025 & 0.053 ± 0.020    & 0.101 ± 0.026    & 0.055 ± 0.149     & 0.534       \\
                                         & 20    & 0.100 ± 0.025 & 0.057 ± 0.036    & 0.100 ± 0.026    & 0.055 ± 0.149     & 0.528       \\
                                         & 21    & 0.100 ± 0.025 & 0.071 ± 0.055    & 0.100 ± 0.026    & 0.055 ± 0.149     & 0.648       \\
                                         & 22    & 0.100 ± 0.025 & 0.056 ± 0.041    & 0.100 ± 0.026    & 0.055 ± 0.149     & 0.681       \\
                                         & 23    & 0.100 ± 0.025 & 0.046 ± 0.035    & 0.101 ± 0.026    & 0.055 ± 0.149     & 0.823       \\
                                         \midrule
\multirow{6}{*}{DistilBERT} & 0     & 0.099 ± 0.026 & 0.086 ± 0.020    & 0.099 ± 0.026    & 0.017 ± 0.069     & -0.133      \\
                                         & 1     & 0.099 ± 0.026 & 0.079 ± 0.019    & 0.099 ± 0.026    & 0.017 ± 0.069     & 0.489       \\
                                         & 2     & 0.099 ± 0.026 & 0.046 ± 0.033    & 0.100 ± 0.026    & 0.017 ± 0.069     & 0.567       \\
                                         & 3     & 0.099 ± 0.026 & 0.044 ± 0.047    & 0.100 ± 0.027    & 0.017 ± 0.069     & 0.594       \\
                                         & 4     & 0.099 ± 0.026 & 0.045 ± 0.035    & 0.100 ± 0.026    & 0.017 ± 0.069     & 0.688       \\
                                         & 5     & 0.099 ± 0.026 & 0.035 ± 0.033    & 0.101 ± 0.026    & 0.017 ± 0.069     & 0.558       \\
\bottomrule
\end{tabular}
\end{table}

\end{document}